\documentclass[lettersize,journal,twoside]{IEEEtran}

\usepackage{times}
\usepackage{soul}
\usepackage{url}
\usepackage[hidelinks]{hyperref}
\usepackage[utf8]{inputenc}
\usepackage{graphicx}
\usepackage{makecell}
\usepackage{amsmath}
\usepackage{booktabs}
\usepackage{hyperref}
\hypersetup{hidelinks,
	colorlinks=true,
	pdfstartview=Fit,
	breaklinks=true}
\usepackage[switch]{lineno}
\usepackage[table]{xcolor}
\usepackage{indentfirst}
\usepackage{epsfig}
\usepackage{amssymb}

\usepackage{multirow}
\usepackage{mathrsfs}
\usepackage{colortbl}
\usepackage{bbding}
\usepackage{doi}

\usepackage[marginal]{footmisc}
\makeatletter

\newcommand{\Rmnum}[1]{\expandafter\@slowromancap\romannumeral #1@}
\makeatother

\begin{document}

% \title{Removal and Selection: Improving RGB-Infrared Object Detection via Coarse-to-Fine Fusion}
\title{Removal then Selection: A Coarse-to-Fine Fusion Perspective for RGB-Infrared Object Detection}

\author{Tianyi~Zhao$^\dag$
, Maoxun~Yuan$^\dag$
, Feng~Jiang, Nan~Wang, Xingxing~Wei$^*$

% <-this % stops a space

\thanks{\quad Corresponding author$^*$: Xingxing Wei.}
\thanks{\quad Xingxing Wei is with the Institute of Artificial Intelligence, Hangzhou Innovation Institute, Beihang University, Beijing, 100191, China (e-mail: xxwei@buaa.edu.cn).}
\thanks{\quad Tianyi~Zhao is with the Institute of Artificial Intelligence, Beihang University, No.37, Xueyuan Road, Haidian District, Beijing, 100191, China (e-mail: ty\_zhao@buaa.edu.cn).}
\thanks{\quad Maoxun~Yuan is with the School of Computer Science and Engineering, Beihang University, No.37, Xueyuan Road, Haidian District, Beijing, 100191, China (e-mail: yuanmaoxun@buaa.edu.cn).}
\thanks{\quad Feng~Jiang and Nan~Wang are with the Beijing Institute of Control and Electronic Technology, Beijing, 100038, China.}
\thanks{\quad$\dag$ represents the equal contribution to this work.}
% <-this % stops a space
}% <-this % stops a space

% \markboth{IEEE Transactions on Multimedia}%
% {ZHAO \MakeLowercase{\textit{et al.}}: A Coarse-to-Fine Fusion Perspective for RGB-Infrared Object Detection}

% \IEEEpubid{0000--0000/00\$00.00~\copyright~2024 IEEE}

\maketitle
\begin{abstract}
In recent years, object detection utilizing both visible (RGB) and thermal infrared (IR) imagery has garnered extensive attention and has been widely implemented across a diverse array of fields. 
By leveraging the complementary properties between RGB and IR images, the object detection task can achieve reliable and robust object localization across a variety of lighting conditions, from daytime to nighttime environments.
Most existing multi-modal object detection methods directly input the RGB and IR images into deep neural networks, resulting in inferior detection performance. We believe that this issue arises not only from the challenges associated with effectively integrating multimodal information but also from the presence of redundant features in both the RGB and IR modalities. The redundant information of each modality will exacerbates the fusion imprecision problems during propagation. To address this issue, we draw inspiration from the human brain's mechanism for processing multimodal information and propose a novel coarse-to-fine perspective to purify and fuse features from both modalities.
Specifically, following this perspective, we design a Redundant Spectrum Removal module to remove interfering information within each modality coarsely and a Dynamic Feature Selection module to finely select the desired features for feature fusion. To verify the effectiveness of the coarse-to-fine fusion strategy, we construct a new object detector called the Removal then Selection Detector (RSDet). Extensive experiments on three RGB-IR object detection datasets verify the superior performance of our method. The source code and results are available at \href{https://github.com/Zhao-Tian-yi/RSDet.git}{https://github.com/Zhao-Tian-yi/RSDet.git}
\end{abstract}


\section{Introduction}
\label{sec:intro}
% Object detection 是计算机视觉的基础任务之一，近些年受到了极大的关注，应用在了监控，遥感，自动驾驶等多个领域。But，研究人员们发现，以往的基于可见光模态的目标检测并不鲁棒，很容易受到环境等因素的影响，因此结合多种传感器模态的融合检测进入到大家的视野，

Object detection is one of the fundamental tasks in computer vision, attracting substantial attention and finding applications in a wide range of fields such as surveillance~\cite{Nascimento2006performance}, remote sensing~\cite{bo2021ship, xia2018dota,yan2022antijamming}, autonomous driving~\cite{wei2023surroundocc, yu2020bdd100k}, etc. However, relying solely on visible imagery for object detection has been shown to be susceptible to various challenges~\cite{wei2023boosting}, like limited illumination, similar appearance of background and foreground, adversarial attack, etc. With the development of sensor technology, various modality images are collected and applied in more and more application fields
% ~\cite{bao2024quality,tu2021multi,tu2022weakly}. 
Therefore, the multi-modal fusion methods~\cite{zhao2021Learning,song2022deep,li2022confidence,tu2023rgbt}
% ~\cite{liu2023deep,wu2023hidanet} 
have come into view. Among them, visible (RGB) and infrared (IR) sensors are widely utilized due to their complementary imaging characteristics. 
Specifically, IR images can clearly provide the outline of the object under poor illumination conditions through the temperature-attached thermal radiation information of the object, which can be regarded as complementary information for RGB images.
Thus, in recent years, researchers have focused on the RGB and IR feature-level fusion ~\cite{yuan2022translation,yuan2023mathbf}, which can help achieve better performance on downstream tasks (e.g., object detection).

\begin{figure}[!t]
    \centering
    \includegraphics[width=\columnwidth]{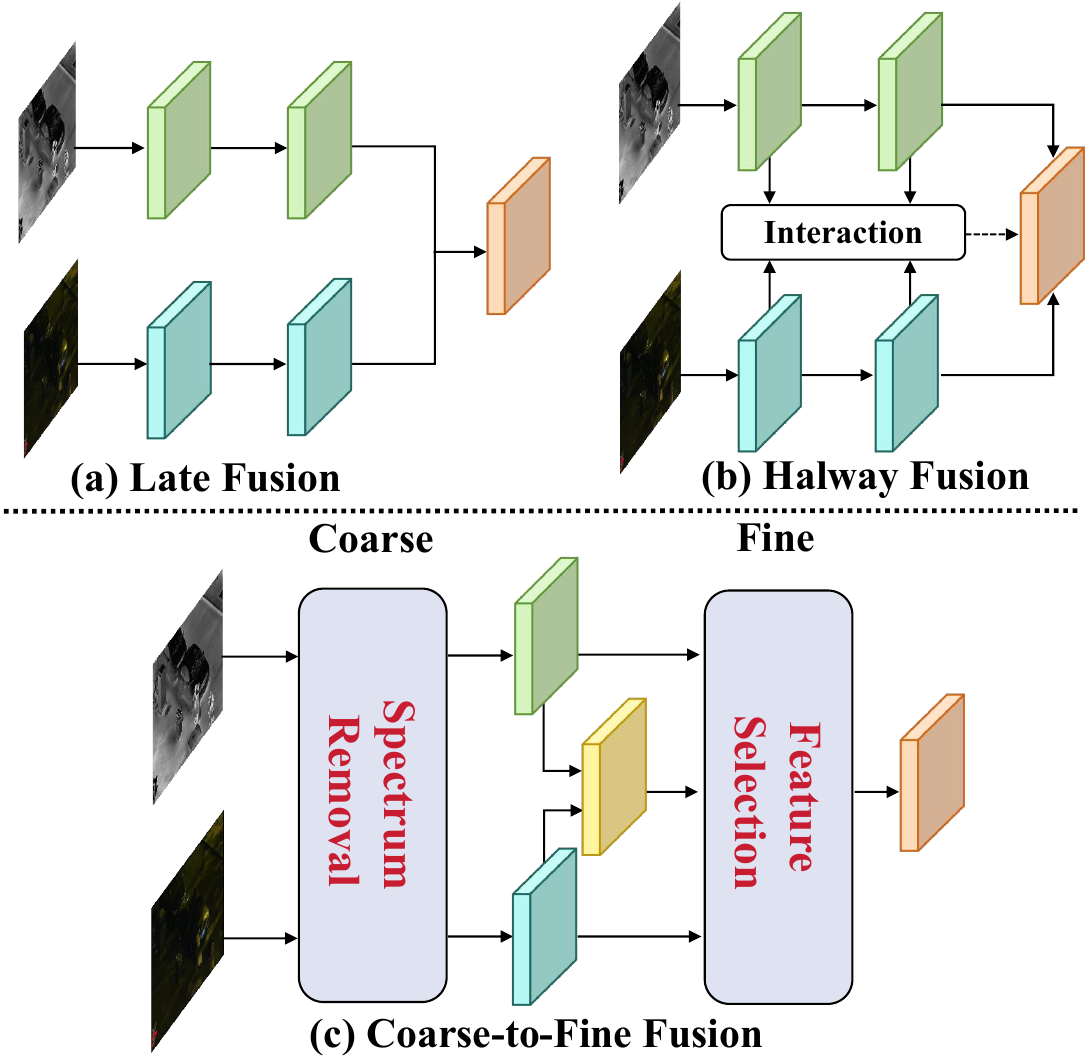}
     % \vspace{-0.2cm}
    \caption{Comparison between existing RGB-IR feature fusion structure and our proposed framework.}
    \vspace{-0.5em}
    \label{fig1}
    % \vspace{-1.5em}
\end{figure}

\begin{figure*}[!t]
    \centering
    \includegraphics[width=0.9\linewidth]{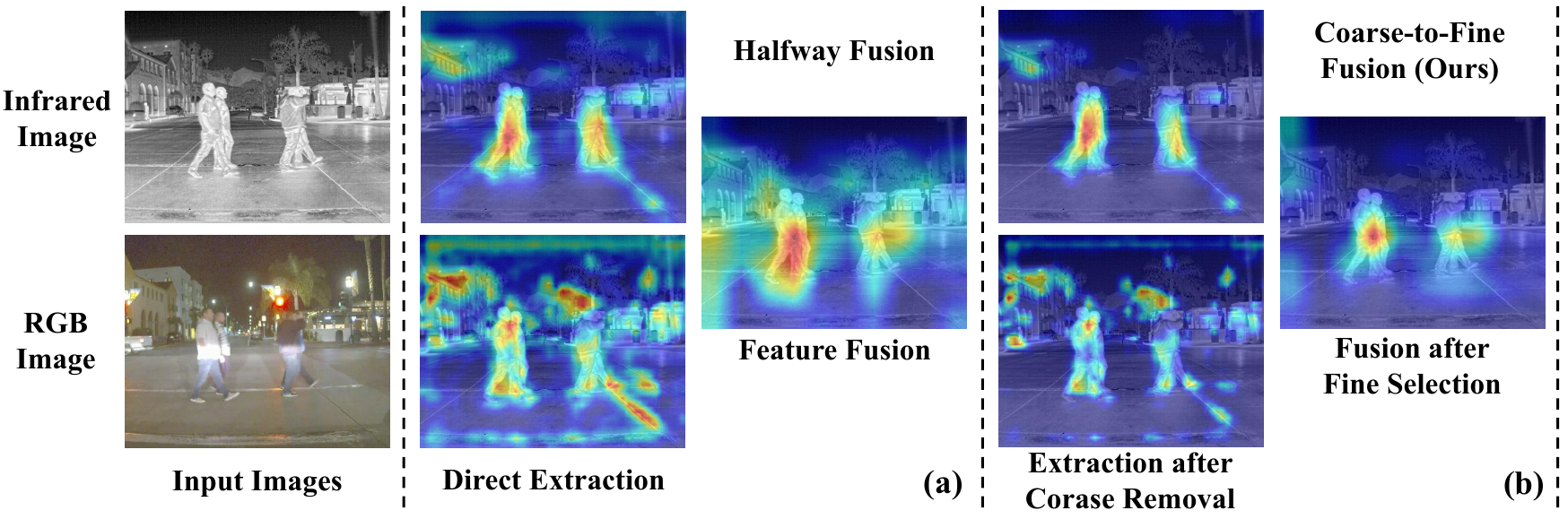}
     % \vspace{-0.2em}
    \caption{Effectiveness of our Coarse-to-Fine fusion. (a) is the current Halfway Fusion method, the directly extracted features are interfered with by the background information from the RGB image and suppress the final fused features, which will result in inferior detection results. (b) Our coarse-to-fine fusion can reduce the irrelevant information and select desired features for fusion, which achieves superior performance.}
    % \vspace{-1.5em}
    \label{fig:2}
\end{figure*}

In RGB-IR object detection, an effective feature fusion method of RGB and IR images is crucial. Most existing RGB-IR object detection methods extract the modality-specific features from RGB and IR images independently, and then directly perform addition or concatenation operations on these features~\cite {liu2016multispectral,li2018multispectral,cao2019box}, as shown in Figure~\ref{fig1}(a). Without explicit cross-modal fusion, the ``Late fusion" strategy is therefore limited in learning the complementary information, resulting in inferior performance. To further explore the optimal fusion strategies, many researchers have explored the ``Halfway fusion" strategy to design an interaction module between different modality features~\cite{zhou2020improving,xie2022learning,li2023multiscale}, as shown in Figure~\ref{fig1}(b). For instance, Zhou \emph{et al.}~\cite{zhou2020improving} construct the MBNet to tap the difference between RGB and IR modalities which brings more useful information at the feature level. Xie \emph{et al.}~\cite{xie2022learning} introduce a novel dynamic cross-modal module that aggregates local and global features from RGB and IR modalities, etc. Although these methods have achieved encouraging improvements, they explicitly enforce the complementary information learning and ignore the negative impact of redundancy features along with the propagation, which would difficult to achieve complementary fusion.

Actually, when confronted with multi-modal information, our brains initially establish rules to filter out interfering information and then meticulously select the desired information, a process that has been modeled in cognitive theory (``Attenuation Theory" \cite{treisman1964selective}). This theory can be likened to a coarse-to-fine process, inspiring us to introduce a new perspective for fusing RGB and IR features. As shown in Figure~\ref{fig1}(c), we design a new fusion strategy called ``Coarse-to-Fine Fusion" to achieve complementary feature fusion. 
``Coarse" indicates that our method begins with the filter out the interfering information and thus can \textbf{coarsely remove} the irrelevant spectrum. To this end, since the redundant information in an image also exists in its frequency spectrum~\cite{li2015finding}, we propose a Redundant Spectrum Removal (RSR) module to filter out coarsely in the frequency domain. Specifically, we convert each source image into frequency space and introduce a dynamic filter to adaptively reduce irrelevant spectrum within RGB and IR modalities. As for ``Fine", it indicates that our fusion strategy conducts \textbf{finely select} features after the coarse removal. We design a Dynamic Feature Selection (DFS) module to meticulously select the desired features between RGB and IR modalities. Therefore, we can weight different scale features required for object detection by exploring a mixture of scale-aware experts. Figure~\ref{fig:2} visualizes an example results of our Coarse-to-Fine fusion strategy. To evaluate the effectiveness of the coarse-to-fine strategy, we construct a novel framework that embeds our coarse-to-fine fusion for RGB-IR object detection called \textbf{R}emoval then \textbf{S}election \textbf{Det}ector~(\textbf{RSDet}).

In summary, this paper has the following contributions:
\begin{itemize}
    \item We propose a new coarse-to-fine perspective to fuse RGB and IR features. Inspired by the mechanism of the human brain processing multimodal information, we coarsely remove the interfering information and finely select desired features for fusion.

    \item Following the coarse-to-fine fusion perspective, we propose a Redundant Spectrum Removal module, which introduces a dynamic spectrum filter to adaptively reduce irrelevant information in the frequency domain. We also design a Dynamic Feature Selection module, which utilizes a mixture of scale-aware experts to weigh different scale features for the RGB-IR feature fusion.

    \item To verify the effectiveness of the coarse-to-fine fusion strategy, we build a novel framework for RGB-IR object detection. Extensive experiments on three public RGB-IR object detection datasets demonstrate our proposed method achieves state-of-the-art performance.
\end{itemize}

The rest of this paper is organized as follows: In Section~\ref{sec: Related Works}, we briefly introduce an overview of relevant studies in RGB-IR Object detection, Shared-Specific Representation learning, and Mixture of Experts. The details of our proposed method are discussed in Section~\ref{sec: Method} and several experiments are conducted to validate the proposed model in Section~\ref{sec: Experiments}. Finally, in Section~\ref{sec: Conclusion}, a conclusion is made for this paper.

\begin{figure*}[!t]
    \centering
    \includegraphics[width=1\linewidth]{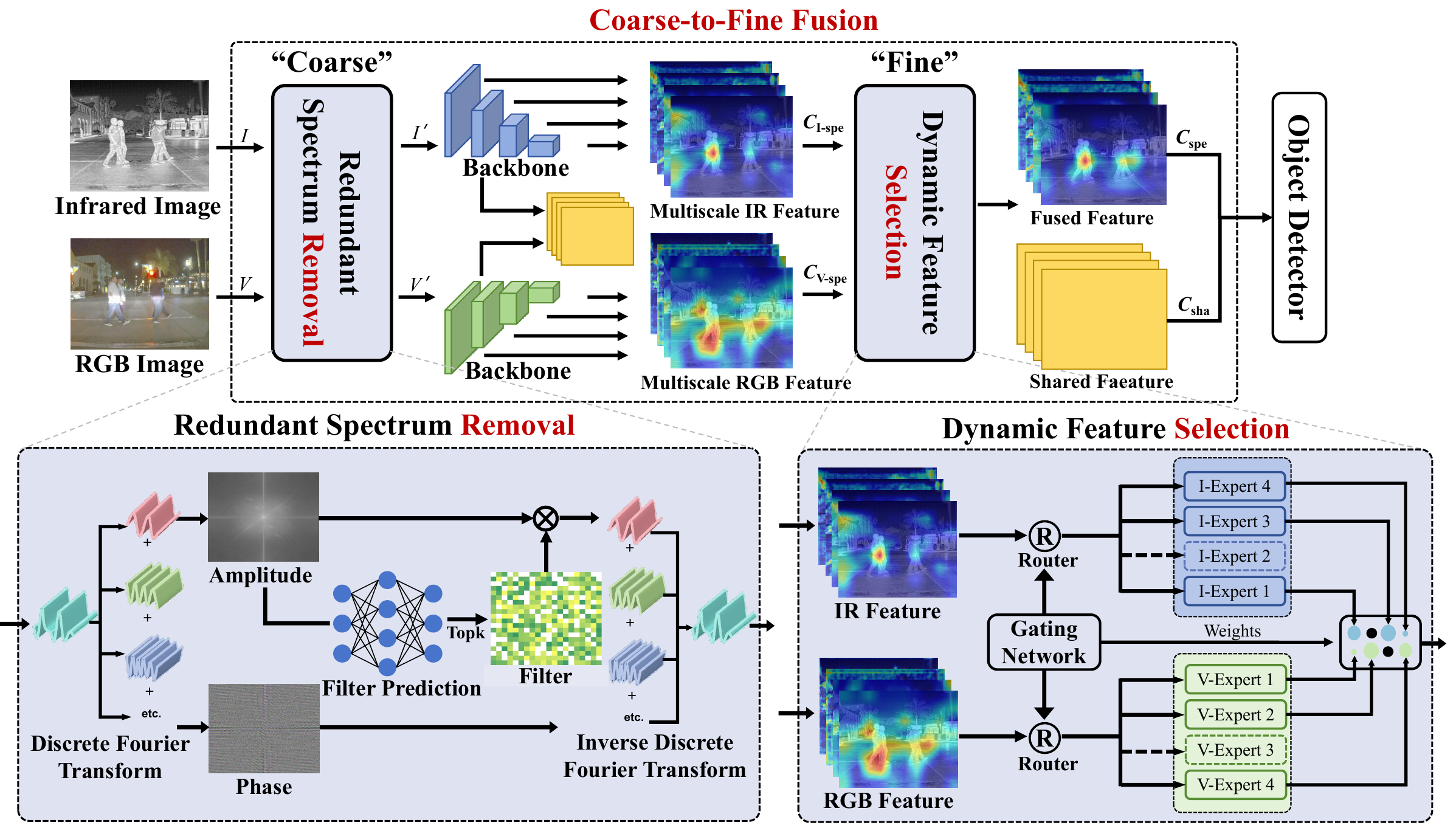}
        % \vspace{0.1em}
    \caption{The overall framework of coarse-to-fine fusion strategy, which mainly consists of the Redundant Spectrum Removal and the Dynamic Feature Selection module. Based on this fusion strategy, a complete object detector named Remove and Select Detector~(RSDet) is constructed to evaluate its effectiveness.}
    % 在caption里面解释清楚coares to fine和RSDet的关系，RSDet不应该出线过早，因为方法中没有提出，先说coares to fine fusion
    % blue}{The overall framework of Coarse-to-Fine Fusion~(above) and the details of the Redundant Spectrum Removal module and the Dynamic Feature Selection module~(below). These components combine with a detector constituting our proposed object detection method, the Removal and Selection Detector~(RSDet).
    % \vspace{-1.5em}
    \label{fig:frame}
\end{figure*}

\section{Related Works} \label{sec: Related Works}
\subsection{RGB-IR Object Detection}
In recent years, thanks to the development of deep learning technology and several visible and infrared datasets being proposed~\cite{hwang2015multispectral,yuan2024improving}, the RGB-IR object detection~(also known as multispectral object detection) task has gradually attracted more and more attention. To fully explore the effective information between visible and infrared images, some researchers focus on the complementarity between the two modalities starting from the illumination conditions. Guan~\emph{et al.}~\cite{guan2019fusion} and Li~\emph{et al.}~\cite{li2019illumination} first propose the illumination-aware modules to allow the object detectors to adjust the fusion weight based on the predicted illumination conditions. Moreover, Zhou ~\emph{et al.} \cite{zhou2020improving} analyze and address the modality imbalance problems by designing two feature fusion modules called DMAF and IAFA. Recently, an MSR memory module~\cite{kim2022towards} was introduced to enhance the visual representation of the single modality features by recalling the RGB-IR modality features, which enables the detector to encode more discriminative features. Yuan~\emph{et al.} \cite{yuan2024improving} propose a transformer-based RGB-IR object detector to further improve the object detector performance. 

In this paper, we draw inspiration from the Attenuation Theory~\cite{treisman1964selective} to emulate how the human brain processes information from multiple sources and propose a coarse-to-fine fusion perspective to utilize RGB and IR features for object detection, which enables the fused features to be more discriminative. Meanwhile, our design of the fusion module ensures the integration effect while also guaranteeing lightweight compared to other fusion modules, as well as lower computational consumption.

\subsection{Shared-Specific Representation Learning}
Shared and specific representation learning is first explored in the Domain Separation Network~\cite{bousmalis2016domain} for unsupervised domain adaptation. It uses a shared-weight encoder to capture shared features and a private encoder to capture domain-specific features. Sanchez~\emph{et al.}~\cite{sanchez2020learning} explored further shared and specific feature disentanglement representation, and found it is useful to perform downstream tasks such as image classification and retrieval. Recently, van~\emph{et al.}~\cite{van2023aspnet} improved the performance of action segmentation by disentangling the latent features into shared and modality-specific components. Furthermore, Wang~\emph{et al.}~\cite{wang2023multi} proposed the ShaSpec model handled missing modalities problems. Shared-specific representation learning has shown great performance and effectiveness in feature learning. However, few RGB-IR object detection models explicitly exploit shared-specific representation. In this paper, we introduce shared-specific representation learning between RGB and IR modality features to implement our coarse-to-fine fusion strategy.

\subsection{Mixture of Experts}

The Mixture-of-Experts (MoE) model \cite{jacobs1991adaptive, jordan1994hierarchical} has demonstrated the ability to dynamically adapt its structure based on varying input conditions. Several studies have been dedicated to the theoretical exploration of MoE~\cite{lepikhin2021gshard}, focusing on the sparsity, training effectiveness, router mechanisms, enhancing model quality, etc. Besides, some researchers also concentrated on leveraging the MoE model for specific downstream tasks. For example, Gross~\emph{et al.}~\cite{gross2017hard} observed that a hard mixture-of-experts model can be efficiently trained to good effect on large-scale multilabel prediction tasks. Cao~\emph{et al.}~\cite{cao2023multi} proposed a mixture of local-to-global experts (MoE-Fusion) mechanisms by integrating MoE structure into image fusion tasks. Chen~\emph{et al.}~\cite{chen2023mod} addressed the multi-task learning by implementing a cooperative and specialized mechanism among experts.

In our proposed method, we introduce the MoE model into the RGB-IR object detection task and propose a mixture of scale-aware experts. Specifically, we design multi-scale experts and leverage its dynamic fusion mechanism to select the desired scale-specific features.

\begin{figure}[!t]
    \centering
    \includegraphics[width=\linewidth]{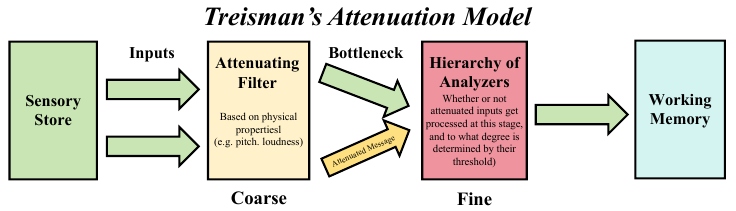}
        % \vspace{-0.5em}
    \caption{Illustration of Treisman's Attenuation Model~\cite{treisman1964selective}}
    % \vspace{-1.5em}
    \label{fig:attention model}
\end{figure}

\section{The Proposed Method} \label{sec: Method}
% In this section, we initially present a novel coarse-to-fine fusion strategy inspired by the Selective Attention Theory in cognitive psychology. Subsequently, we introduce the Redundant Spectrum Removal (RSR) module, designed to filter out task-irrelevant information in the frequency domain of RGB and IR images. Following that, the Dynamic Feature Selection (DFS) module is proposed, utilizing a series of scale-aware experts to selectively integrate features across different scales. Ultimately, these modules are integrated into a shared-specific network architecture to facilitate the complementary fusion of RGB and IR features, resulting in the Removal and Selection Detector (RSDet).

% In the next sections, we will further introduce the proposed RSR module and DFS module.
\subsection{``Coarse-to-Fine" Fusion} \label{subsec: RSDet}
The proposed ``Coarse-to-Fine" Fusion strategy is inspired by cognitive models of human information processing, specifically the Selective Attention Theory in cognitive psychology. Notable examples include Broadbent's Filter Model~\cite{broadbent1958perception} and Treisman's Attenuation Model~\cite{treisman1964selective}. These models serve as a cornerstone of attention mechanism theory in cognitive psychology. As shown in Figure~\ref{fig:attention model}, Treisman's Attenuation Model posits that when the human brain processes multiple stimuli, it first attenuates unimportant or irrelevant messages based on specific criteria. Then it processes the remaining messages in a more refined manner which conducts detailed, hierarchical analysis and processing to extract meaningful features and insights.
Finally, the processed messages enter the brain's working memory.

% % 下面这段应该主要讲coarse2fine fusion ，而不应该出现具体模块的名称
% \textcolor{red}{As illustrated in Figure~\ref{fig:attention model}, the Attenuating Filter in Treisman's Attenuation Model inspires the Redundant Spectrum Removal (RSR) module, which filters out irrelevant or interfering information based on specific criteria or rules when dealing with multi-sensory inputs in computer vision systems. Similarly, the Hierarchy of Analyzers in Treisman’s Attenuation Model provide inspiration for our Dynamic Feature Selection (DFS) module, which conducts detailed, hierarchical analysis and processing to extract meaningful features and insights, akin to advanced processing layers in deep learning models. Finally, just as processed information enters the brain's working memory, the refined features from our model are passed to the detection network for final object detection and analysis.}

Inspired by Treisman's Attenuation Model, we design the ``Coarse-to-Fine" Fusion strategy. ``Coarse" corresponds to the Redundant Spectrum Removal (RSR) module to filter out coarsely in the frequency domain, and `Fine' corresponds to the Dynamic Feature Selection (DFS) module to meticulously select the desired features between RGB-IR modalities. 
Since the two modality features often intersect, we introduce disentangled representation learning~\cite{sanchez2020learning} to purify and decouple them for complementary fusion. As shown in Figure~\ref{fig:frame}, we integrate the RSR and DFS modules into shared-specific structures to implement the ``Coarse-to-Fine" Fusion. Firstly, input the RGB~($V$) and IR~($I$) images to the RSR module separately, removing interfering information to obtain the images~$V^{'}$ and~$I^{'}$ with the irrelevant redundant spectra removed. Then, we introduce the shared-specific structure to extract the two modality-specific multi-scale features $C_{\text{I-spe}}$ and $C_{\text{V-spe}}$, which uses ResNet as the backbone network. As for the shared features $C_{\text{sha}}$, we also employ several Resblocks as the feature extractor. After that, these different scale features $C_{\text{I-spe}}$ and $C_{\text{V-spe}}$ are input to the DFS module, which can dynamically aggregate them by the proposed mixture of scale-aware experts and obtain the specific feature $C_{\text{spe}}$. Finally, the specific feature $C_{\text{spe}}$ and the shared feature $C_{\text{sha}}$ are added together to get the final fused feature~$C$, which can be expressed as:
\begin{equation}
    C= C_{\text{sha}} + C_{\text{spe}}.
\end{equation}

\subsection{Redundant Spectrum Removal} \label{subsec: RSR}
% 这里为什么在频域解释过于简单，要点出相比于在空间域上处理，在频域处理的优势
For `Coarse', we choose to process the image in the frequency domain, due to the frequency domain having inherent global modeling properties, and only through positional multiplication operations can filter out features of the same frequency band in the entire image. But, it is difficult to handle the tight coupling of object features in the spatial domain.
Therefore, we propose a Redundant Spectrum Removal (RSR) module to perform coarse filtering in the frequency domain. We first transform each input image into the frequency domain and predict a dynamic filter to attenuate irrelevant spectrum within RGB and IR modalities adaptively. 

\begin{figure}[!t]
    \centering
\includegraphics[width=\linewidth]{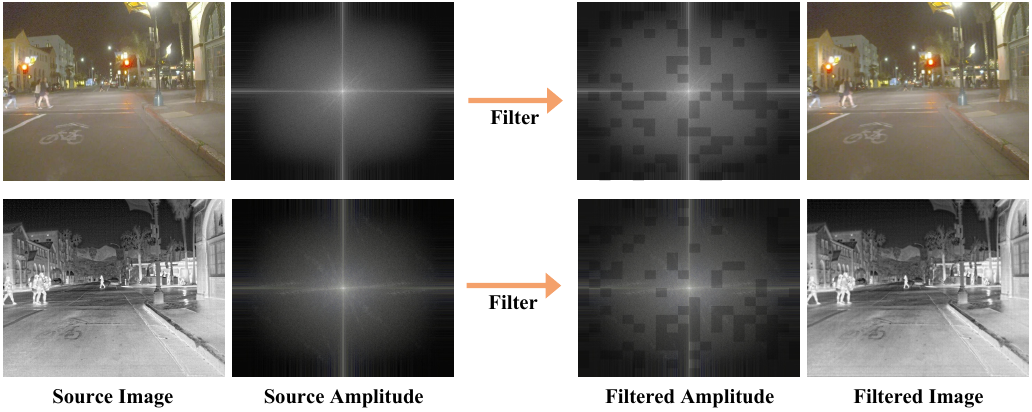}
     \vspace{-0.5em}
    \caption{Example visualization of the learnable filter $H_I(u,v)$ and $H_V(u,v)$.}

    \vspace{-1.0em}
    \label{fig:rsr_filter}
\end{figure}

Specifically, the paired RGB image $V \in \mathbb{R}^{H \times W \times 3}$ and IR image $I \in \mathbb{R}^{H \times W \times 1}$ are taken as the RSR module input. They are subjected to the Discrete Fourier Transform~($\operatorname{DFT}(\cdot)$) and get the frequency domain image $F_I(u,v)$ and $F_V(u,v)$:
\begin{equation}
    \begin{aligned}
       & F_I(u,v) = \operatorname{DFT}(I),\\
       & F_V(u,v) =\operatorname{DFT}(V).& 
    \end{aligned}
\end{equation}
The filter prediction network is designed to dynamically generate the redundant spectrum filter based on the amplitude information $|F_I(u,v)|$ and $|F_V(u,v)|$ of the original images, which is illustrated in Figure \ref{fig:frame}. For each modality image, we perform a simple encoder on the amplitude image to obtain a feature embedding:

\begin{equation}
\begin{aligned}
   &{M}_{l_I} = \operatorname{Encoder}_I(|F_I(u,v)|),\\
   &{M}_{l_V} = \operatorname{Encoder}_V(|F_V(u,v)|).
\end{aligned}
\end{equation}
% 这里是不是只需要给出过程就可以了，不需要给出具体数值吧，比如分成N patches
% 请确认红色部分公式没有问题
Each value of embedding ${M}_{l_I}, {M}_{l_V}$ $\in \mathbb{R}^m$ represents the importance of different patch regions of $F_I(u,v)$ and $F_V(u,v)$ image. Then to fully retain the effective spectrum components while attenuating the useless ones, we utilize the $\operatorname{top}K$ operation on $M_{l_I}$ and $M_{l_V}$:
\begin{equation}
\begin{aligned}
      &M_{m_I} = \operatorname{top}K(M_{l_I}),\\
      &M_{m_V}= \operatorname{top}K(M_{l_V}).
\end{aligned}
\end{equation}
Next, we use nearest neighbor interpolation to reshape the embedding to match the size of the original image, obtaining the filters $H_I(u,v)$ and $H_V(u,v)$: 
\begin{equation}
\begin{aligned}
    & H_I(u,v)= \operatorname{Reshape}(M_{m_I}),\\
    & H_V(u,v) = \operatorname{Reshape}(M_{m_V}).
\end{aligned}
\end{equation}
Intuitively, we visualize the learnable filters $H_I(u,v)$ and $H_V(u,v)$ as shown in Figure~\ref{fig:rsr_filter}. From the 'Filtered Amplitude', we can observe that the learned filters remove some high-frequency noise in each modality, which can be considered as redundant information for the object detection task.

Subsequently, we perform element-wise multiplication of $H_I(u,v)$ and $H_V(u,v)$ with the frequency domain images $F_I(u,v)$ and $F_V(u,v)$:
\begin{equation}
\begin{aligned}
   F_I^{'}(u,v) = F_I(u,v) \otimes H_I(u,v),\\
   F_V^{'}(u,v) = F_V(u,v) \otimes H_V(u,v).
\end{aligned}
\end{equation}

Finally, the filtered $F_I^{'}(u,v)$ and $F_V^{'}(u,v)$ are subjected to the Inverse Discrete Fourier Transform, denoted as $\operatorname{IDFT}(\cdot)$, to transform the images back to the spatial domain. This process yields the images $I^{'}$ and $V^{'}$, with the redundant and irrelevant spectrum removed.
\begin{equation}
\begin{aligned}
    I^{'}= \operatorname{IDFT}(F_I^{'}(u,v)),~V^{'} = \operatorname{IDFT}(F_V^{'}(u,v)).
\end{aligned}
\end{equation}

\begin{figure}[!tbp]
    \centering

    \includegraphics[width=1\linewidth]{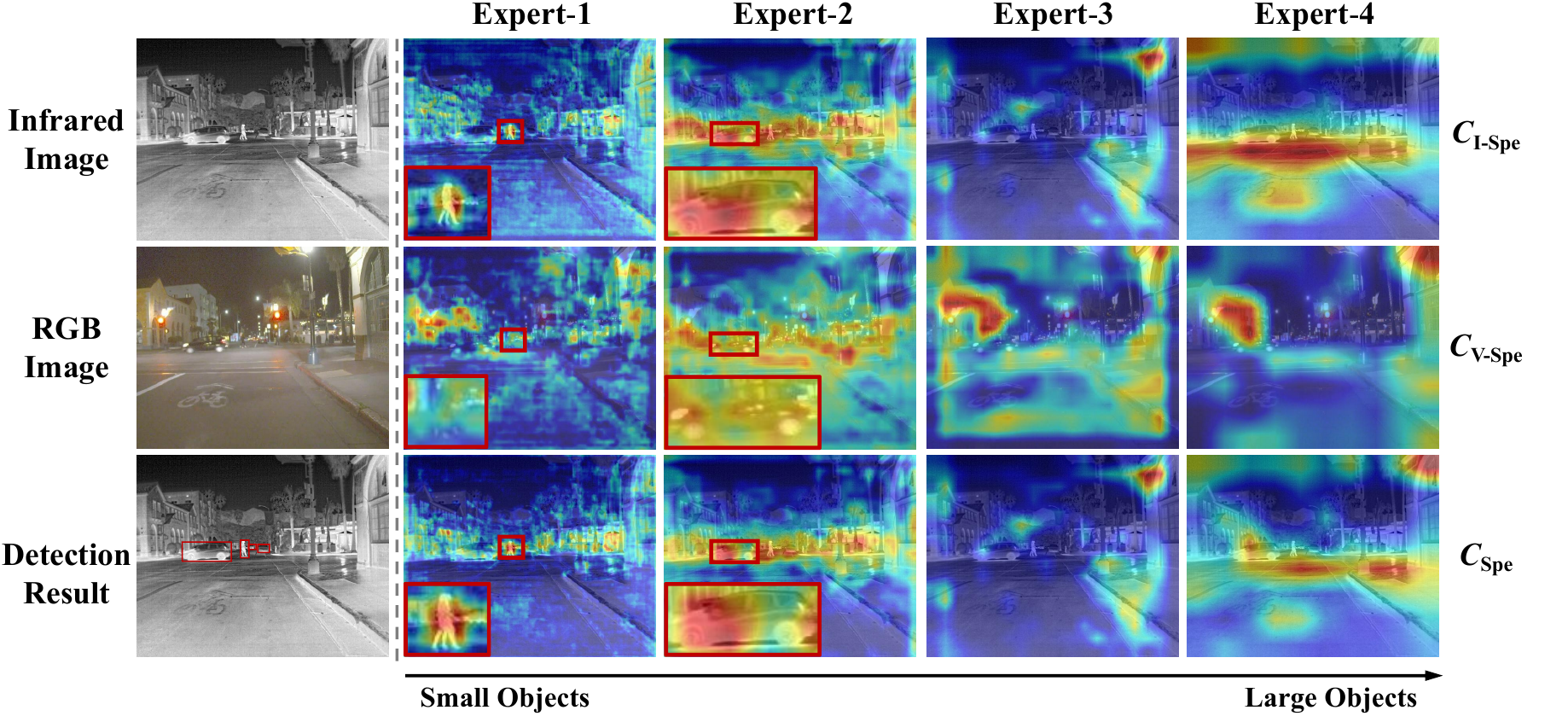}
        \vspace{-0.5em}
    \caption{Visualization of the features learned by each expert ($C_{I-Spe}$ and $C_{V-Spe}$) and the fused feature ($C_{Spe}$). To facilitate a clearer observation, we use red boxes to highlight the objects and overlay the features onto the original RGB or IR image. From left to right the feature scale from large to small, corresponding to the objects size from small to large.}
        \vspace{-1.0em}
    \label{fig:supp_moe_flir}
\end{figure}

\subsection{Dynamic Feature Selection}\label{subsec: DFS}
% why use it

For ``Fine", we implement the dynamic modality feature selection in the fusion process by employing a mixture of scale-aware experts to gate multi-scale features leveraging its dynamic fusion mechanism to facilitate complementary fusion across different scales.
As shown in Figure~\ref{fig:frame}, we design a dedicated expert for each scale modality-specific feature. Then, we aggregate these features by using the gating network to predict a set of dynamic weights. Specifically, after obtaining different scale features $C_{\text{I-spe}}^i$ and $C_{\text{V-spe}}^i$ through the feature extraction for $I^{'}$ and $V^{'}$, we first make $C_{\text{I-spe}}^i$ and $C_{\text{V-spe}}^i$ go through the average pooling operations, and flattened them into one-dimension vector $X_{I}^i\in\mathbb{R}^{M}$ and $X_{V}^i\in\mathbb{R}^{M}$ to predict the weights $w_I^i$ and $w_V^i$ of the gating network $G$. It can be formulated as follows:
\begin{equation}
\begin{aligned}
    w_I^{i}, w_V^{i} = G\left(X_{I}^{i}, X_{V}^{i} \right)=\operatorname{Softmax}\left(\left[X_{I}^{i}, X_{V}^{i}\right]\cdot W_{\text {}}\right), 
\end{aligned}
\end{equation}
where the $W \in \mathbb{R}^{M\times N}$ is the learnable weight matrix normalized through the softmax operation and $i$ is the index of experts.
After that, the weights $w_I$ and $w_V$ are converted to gating via the Router $R$, preserving the desired features between the two modalities for fusion at different scales. Consequently, the Router $R$ can be  formulated as follows:

\begin{equation}
    \begin{gathered}
    (r_I^i,r_V^i)=R(w_I^i, w_V^i)=\begin{cases}
     (1,1), & w_I^i,w_V^i \ge t \\ 
     (1,0), & w_I^i \ge t , w_V^i \le t, \\ 
     (0,1), & w_I^i \le t , w_V^i \ge t 
 \end{cases}
\end{gathered}
\label{eq: router}
\end{equation}
where $t$ is a threshold.
% 时态和 口语化，太粗糙了
% It should be noted that, when calculating the feature weights $w_I^{i}, w_V^{i}$, we were concerned that this approach might result in a modality preference towards one dominant modality~(for example, regardless of the quality of the RGB image, the network tends to use only the infrared modality). Therefore, we \textcolor{red}{think about} add a regularization term between the two weights to prevent the DFS module from showing a strong preference towards a specific modality. But, during the experiments, we found that regardless of whether regularization terms were added, the modality preference phenomenon did not occur. Therefore, we did not include a modality balancing regularization term in our method.}
Then $N$ scale-aware expert networks $\mathcal{E}_{I}^{i}$ and $\mathcal{E}_{V}^{i}$  are utilized to further extract modality-specific features. The formalization is as follows:
\begin{equation}
\begin{aligned}
    & C_I^i= \mathcal{E}_{I}^{i}(x_I^{i}\cdot r_I^i),~C_V^i= \mathcal{E}_{V}^{i}(x_V^{i}\cdot r_V^i),
\end{aligned}
\end{equation}
where $x_I^{i}$ and $x_V^{i}$ are the multiscale features of different modality input  $I^{'}$ and $V^{'}$. The detailed structure of each scale-aware expert $\mathcal{E}$ is the same and mainly consists of two convolution blocks.
After obtaining the output results from expert models at different scales, we perform dynamic weighted summation and concatenate them together to get the ultimate multi-modal specific feature $C_{\text{spe}}$:
\begin{equation}
\begin{aligned}
    C_{\text{spe}}= \bigcup_{i=1}^{n}\left(w_I^i C_I^i+w_V^i C_V^i\right).
\end{aligned}
\end{equation}

To illustrate the effectiveness of the DFS module, we visualise the features extracted by four different experts, as shown in Figure~\ref{fig:supp_moe_flir}. From left to right, it can be seen that different experts focus on different scales objects. 
Each expert selects the desired features from two modality features ($C_{I-Spe}$ and $C_{V-Spe}$) so that the fused features $C_{Spe}$ exhibit greater prominence for object detection.
The visualisation results indicate that the fusion method effectively integrates complementary features from different modalities across different scales, leading to a comprehensive representation.

\subsection{Removal and Selection Detector (RSDet)}  \label{subsec: Loss Function}
% \noindent\textbf{Overall framework.}~
To evaluate the effectiveness of our Coarse-to-Fine fusion strategy, we embed it into an existing object detection framework. In specific, we utilize a two-stage detector Faster R-CNN \cite{ren2017faster} as our baseline model and replace its backbone with our strategy to construct a new object detector called RSDet. Other modules such as Region Proposal Network (RPN) and R-CNN head remain unchanged.

\noindent\textbf{Loss functions.}~To extract the shared and specific feature from the images $I^{'}$ and $V^{'}$ after RSR module, we maximize the mutual information~\cite{tschannen2019mutual} between~$C_{\text{I-spe}}$~and~$C_{\text{V-spe}}$~with~$C_{\text{sha}}$~. The mutual information can serve as the deep supervising loss function $\mathcal{L}_{\text {I-spe}}$ and $\mathcal{L}_{\text {V-spe}}$ to guide the shared-specific features learning. The definitions are as follows:

\begin{equation}
     \mathcal{L}_{\text {I-spe}} = \operatorname{MI}(C_\text{sha},C_\text {I-spe}),
\end{equation}
\begin{equation}     
     \mathcal{L}_{\text {V-spe}} = \operatorname{MI}(C_\text {sha},C_\text {V-spe}),
\end{equation}
where $\operatorname{MI}$ represents the operation of mutual information. We use cross-entropy~($\operatorname{CE}$) and KL-divergence~($\operatorname{KL}$) to approximate equivalent optimize the mutual information between different features in the latent space. 
\begin{equation}
\begin{aligned}
    \max\operatorname{MI}(x,y) \Rightarrow  \max\{\operatorname{CE}(x,y) -\operatorname{KL}(x||y) \\+\operatorname{CE}(y,x) -\operatorname{KL}(y||x)\}.
    % \Longleftrightarrow &\max \left\{ H(P)+H(Q)-H(P,Q) \right\} \nonumber\\
\end{aligned}
\label{eq:MI}
\end{equation}
As for detection loss, we use the $\mathcal{L}_{\text{rpn}}$, $\mathcal{L}_{\text {reg}}$ and~$\mathcal{L}_{\text {cls}}$~same as the Faster R-CNN~\cite{ren2017faster}~to supervise the detection process of RSDet. The total loss is the sum of these individual losses: 
\begin{equation}
    \mathcal{L} = \gamma (\mathcal{L}_{\text {I-spe}}+ \mathcal{L}_{\text {V-spe}})+\mathcal{L}_{\text {rpn}}+\mathcal{L}_{\text {reg}}+ \mathcal{L}_{\text {cls}},
\end{equation}
where $\gamma=0.001$ is the coefficient used to strike a balance between the different loss functions.

\begin{table*}[!t]
    \centering
    \caption{Ablation study on each module result on the FLIR, LLVIP~(mAP, in\%)  and KAIST~($\text{MR}^{\text{-2}}$ in\%) dataset, under IoU=0.7. The best results are highlighted in \textbf{bold}.}
            \vspace{-0.5em}
        % \vspace{-1.0em}
    \renewcommand{\arraystretch}{1.1}
    \setlength{\tabcolsep}{2.5mm}
\begin{tabular}{cc|ccc|ccc|ccc}
    \hline

&  & \multicolumn{3}{c|}{\textbf{FLIR}} & \multicolumn{3}{c|}{\textbf{LLVIP}} &\multicolumn{3}{c}{\textbf{KAIST~(`All' Setting)}}\\ \cline{3-11}
\multirow{-2}{*}{\textbf{RSR}} & \multirow{-2}{*}{\textbf{DFS}} &\textbf{$\text{mAP}_{50}$~$\uparrow$} &\textbf{$\text{mAP}_{75}$~$\uparrow$}& \textbf{mAP~$\uparrow$}&\textbf{$\text{mAP}_{50}$~$\uparrow$} &\textbf{$\text{mAP}_{75}$~$\uparrow$}& \textbf{mAP~$\uparrow$} & \textbf{$\text{MR}^{\text{-2}}$-Day~$\downarrow$}&\textbf{$\text{MR}^{\text{-2}}$-Night~$\downarrow$}& \textbf{$\text{MR}^{\text{-2}}$-All~$\downarrow$} \\\hline
        &        &81.2&36.2&41.2   &95.0&58.6& 55.5 &25.30& 28.87&  26.48  \\

\checkmark&  
&82.3&36.0&41.9&95.0&62.0&57.1&25.27& 27.89&  26.12\\
&\checkmark  &{82.4}&38.7&42.8&95.6&66.0&59.5 &25.11& 27.99&26.23\\
%     &        & 75.2&31.4 &37.2  &95.0&58.6& 55.5 &&&   \\
% \checkmark&  &76.3&34.1& 38.4&95.0&62.0&57.1&&& \\
% &\checkmark  &79.8&34.8&40.8&95.6&66.0&59.5 && &\\
\checkmark        & \checkmark     & \textbf{83.9} &\textbf{40.1}& \textbf{43.8}&\textbf{95.8}&\textbf{70.4}&\textbf{61.3}&\textbf{24.18} &\textbf{26.49}& \textbf{24.79}\\     \hline
\end{tabular}
    % \vspace{-1.0em}
\label{tab:ablation}
\end{table*}

\begin{table*}[!tbp]
\begin{center}
  \caption{Comparsion our DFS module with other feature fusion methods under a fair experiment setting on the FLIR dataset. Uniformly utilize Faster R-CNN as the detector, with ResNet-50 as the backbone.  The best results are highlighted in \textbf{bold}.}
        \vspace{-0.5em}
    \label{tab: fusion method}
    \setlength{\tabcolsep}{3.5mm}
        \renewcommand{\arraystretch}{1.1}
\begin{tabular}{c|c|cc|cc|ccc}
\hline
\textbf{Index}& \textbf{Modality} & \multicolumn{2}{c|}{\textbf{Methods}}   & \textbf{Param.} & \textbf{GFLOPs}&\textbf{$\text{mAP}_{50}$~$\uparrow$ }&\textbf{$\text{mAP}_{75}$~$\uparrow$}& \textbf{mAP~$\uparrow$}  \\ \hline
(a)&RGB      & \multicolumn{2}{c|}{\multirow{2}{*}{Faster R-CNN~\cite{ren2017faster}}}   & 41.13M  & 75.6  & 64.9  & 21.1  & 28.9 \\
(b)&IR  &  &  & 41.13M  & 75.6  & 74.4  & 32.5  & 37.6 \\\cline{3-9}
(c)&RGB+IR     &  \multirow{4}{*}{\makecell[c]{Two Stream \\ Faster R-CNN~\cite{ren2017faster}}}&Two Stream& 64.61M  & 102.5 & 73.1  & 32.0  & 37.1 \\
(d)&RGB+IR     & &+~CMX~\cite{zhang2023cmx}   &305.20M 
        & 229.0    &   80.5   &    33.4     &  39.7    \\
(e)&RGB+IR     & &+~CFT~\cite{qingyun2021cross}      & 196.93M & 136.8 & 77.5  & 34.6 & 39.2 \\
(F)&RGB+IR     & &+~DFS~(Ours)  &   68.52M      &   136.0     & \textbf{80.9} & \textbf{36.0}&\textbf{41.3}\\\hline

\end{tabular}
        \vspace{-1em}
\end{center}
\end{table*}

\begin{table}[!tbp]
\centering
  \caption{
  comparison of different design choices for the filter Type and the $K$ value of the Topk operation in the RSR module on the FLIR dataset. The best results are highlighted in \textbf{bold}.}
        \vspace{-0.5em}
    \label{tab: design rsr}
    \setlength{\tabcolsep}{3mm}
        \renewcommand{\arraystretch}{1.1}
\begin{tabular}{c|c|ccc}
\hline
\textbf{Filter Type} &\textbf{K}  &   \textbf{$\text{mAP}_{50}$~$\uparrow$} &\textbf{$\text{mAP}_{75}$~$\uparrow$}& \textbf{mAP~$\uparrow$}  \\ \hline
\multirow{7}{*}{Hard Filter}&300&82.7 &36.4 &42.0\\
&320&83.2&38.4&{43.3} \\
&340& 80.8&36.0 & 41.6\\
&360& 81.9&36.9 &42.1 \\
&380& 83.5& {38.8}&{43.3} \\
&400& 82.4 &38.7 &42.8 \\\cline{2-5}
& Avg.&82.42&37.53& 42.52\\\hline
\multirow{7}{*}{Soft Filter}&300& {83.8}& {38.2}& {43.2}\\

&320& \textbf{83.9}&\textbf{40.1} &\textbf{43.8} \\
&340& 82.6& {38.8}&43.1 \\
&360& 82.2&37.7 &42.6 \\
&380& {83.5}&36.6 & 42.7\\
&400& 82.4 &38.7 &42.8  \\\cline{2-5}
& Avg.&82.90&38.17&42.90 \\\hline
\end{tabular}
        \vspace{-1.0em}
\end{table}

\section{Experiments} \label{sec: Experiments}

\subsection{Experimental Setup}
\subsubsection{\textbf{Datasets}}
\ 
\newline
\indent\textbf{KAIST}~\cite{hwang2015multispectral} is a public multispectral pedestrian detection dataset. Due to the problematic annotations in the original dataset, further research has improved the annotations of train~\cite{zhang2019weakly} and test dataset~\cite{liu2016multispectral}. Our method is trained on 8,963 image pairs and evaluated on 2,252 image pairs with the improved annotations. The KAIST dataset is divided into different subsets~\cite{hwang2015multispectral}: near, medium, and far~(``Scale"); none, partial, and heavy~(``Occlusion"); day, night, and all"~(All and Reasonable). In particular, the `All' setting evaluates the model performance on all objects of the KAIST test dataset, while the `Reasonable' setting only consists of not/partially occluded objects and objects larger than 55 pixels. 
To comprehensively evaluate the performance of our method, we perform comparison experiments under `All' settings.

\textbf{FLIR-aligned} is a paired RGB-IR object detection dataset including daytime and night scenes. Since the images are misaligned in the original dataset, we use the FLIR-aligned dataset~\cite{zhang2020multispectral}. It has 5,142 aligned RGB-IR image pairs, of which 4,129 are used for training and 1,013 for testing, and contains three classes of objects: 'person', 'car', and 'bicycle'. Since there are very few instances of the 'dog' category in the FLIR-aligned dataset, we clean the annotations and remove the 'dog' category from the dataset.

\textbf{LLVIP}~\cite{jia2021llvip} is a strictly aligned RGB-IR object detection dataset for low-light vision. It is collected in low-light environments, and most of the data are captured in very dark scenes. It contains 15,488 aligned RGB-IR image pairs, of which 12,025 images are used for training and 3463 images for testing.

\subsubsection{\textbf{Evaluation Metrics}}
\ 
\newline
\indent\textbf{Log-average Miss Rate~(\texorpdfstring{$\text{MR}^{\text{-2}}$})):} For the KAIST dataset, we employ the standard KAIST evaluation \cite{hwang2015multispectral}: Miss Rate (MR) over False Positive Per Image~(also denoted as $\text{MR}^{\text{-2}}$). It calculates the average miss rate under the 9 FPPI values which are sampled uniformly in the logarithmic interval. The lower values indicate better performance.

\textbf{mean Average Precision~(mAP):} For FLIR and LLVIP datasets, we employ the commonly used object detection metric Average Precision (AP). The positive and negative samples should be divided according to the correctness of classification and Intersection over the Union (IoU) threshold. 
The $\text{mAP}_{50}$ metric represents the mean AP under IoU=0.50 and the mAP metric represents the mean AP under IoU ranges from 0.50 to 0.95 with a stride of 0.05. 

\begin{figure*}[!t]
    \centering
\includegraphics[width=\linewidth]{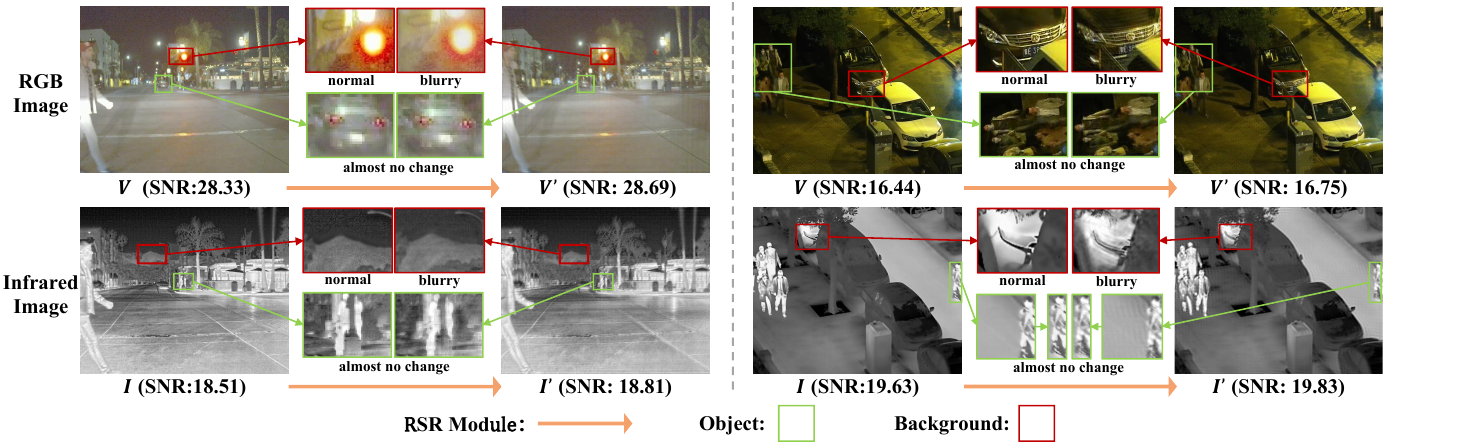}
     \vspace{-1.5em}
    \caption{Visualization of RSR module intermediate output on the FLIR~(left) and LLVIP~(right) datasets.
     $V$ and $I$ represent the input RGB and Infrared images respectively, $V^{'}$ and $I^{'}$ represent the images with irrelevant redundant spectrum removed after the RSR module. ``SNR" stands for signal-to-noise ratio. The green bounding box indicates the object~(foreground), while the red bounding box indicates the background~(no-object class). We have provided enlarged views of these regions to facilitate observing and comparing the changes in the object and background regions before and after the RSR module.
     }
    % \vspace{-1.0em}
    \label{fig:rsr}
\end{figure*}

\begin{figure*}[!t]
    \centering
    \includegraphics[width=0.9\linewidth]{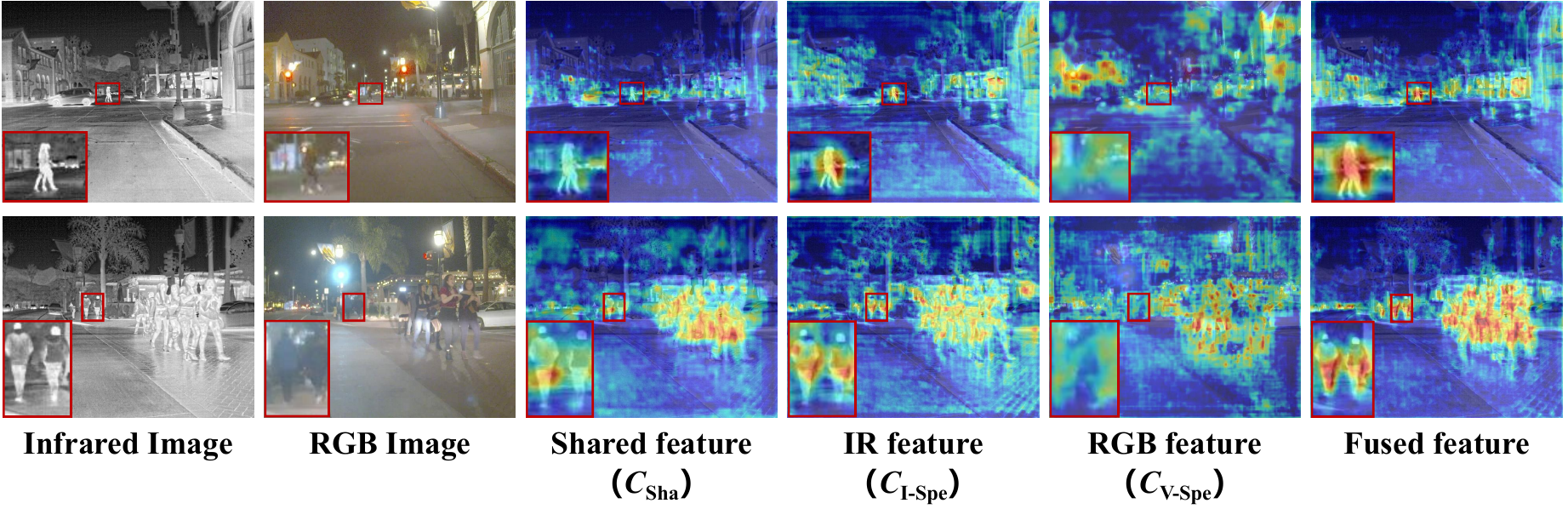}
          \vspace{-0.5em}
    \caption{Visualization of DFS module feature fusion results on the FLIR dataset. To facilitate a clearer observation, we overlaid the features onto the original RGB or Infrared image. }
    % \vspace{-0.5em}
    % \label{fig2}
    % \vspace{-2em}
      \vspace{-1.0em}
    \label{fig: DFS_vis}
\end{figure*}

\begin{figure}[!t]
    \centering
    \includegraphics[width=0.9\linewidth]{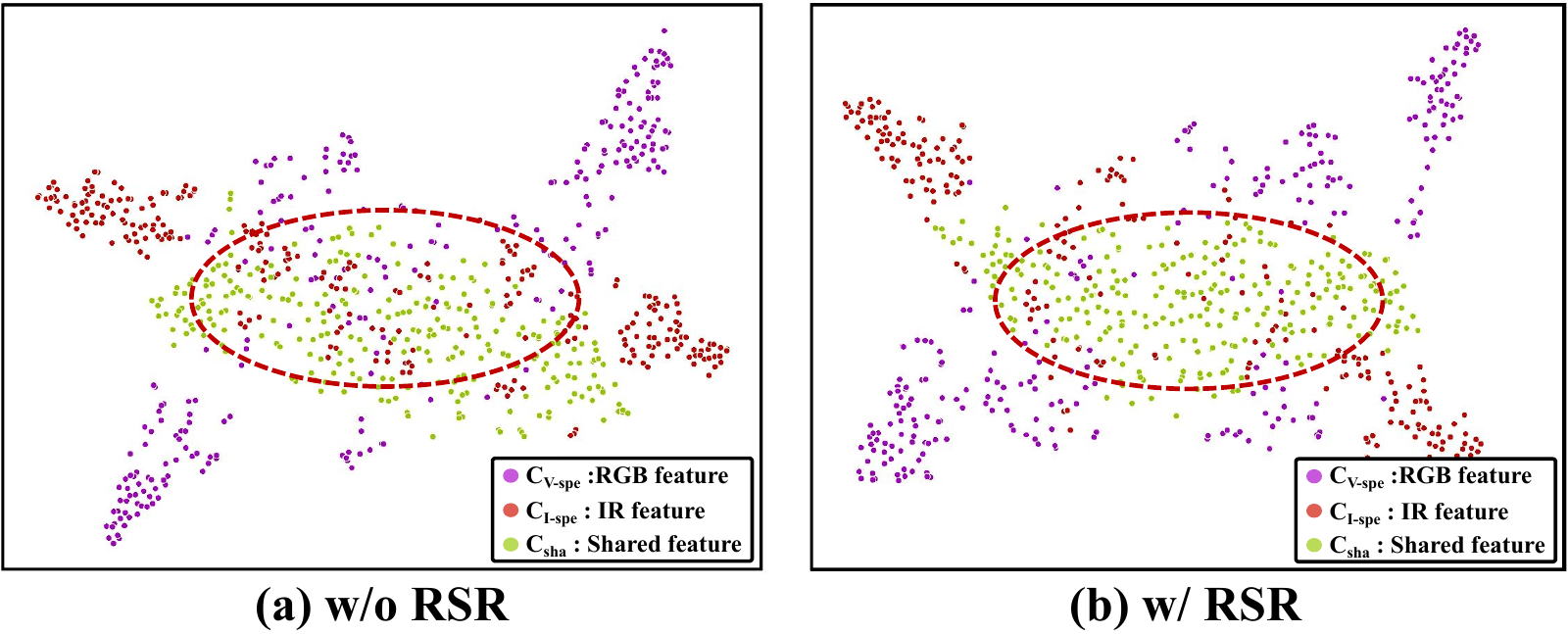}
        % \vspace{-0.5em}
    \caption {tSNE visualization of the modality-specific and shared features. `w/o RSR'(a) and `w RSR' (b) represent without and with RSR module.}
    \vspace{-1.2em}
    % \vspace{-1em}
    \label{fig: rsr_tsne}
\end{figure}

\subsubsection{\textbf{Implementation Details}}
\ 
\newline
All the experiments are implemented in the mmdetection toolbox and conduct on the NVIDIA GeForce RTX 3090. We use the Faster R-CNN as the object detector with ResNet50 as the backbone. During the training phase, the SGD optimizer is employed with a momentum of 0.9 and a weight decay of $1\times10^{-4}$. To facilitate the design of the DFS module, we adjusted the input image resolution of various datasets to match that of the LLVIP dataset.
All experiments are most trained for 12 epochs with an initial learning rate of $1\times10^{-2}$ for the FLIR-align and KAIST datasets, and $1\times10^{-3}$ for the LLVIP datasets. As for data augmentation, we only use random flipping with a probability of 0.5 to increase input diversity. 

% To facilitate the design of the DFS module, we adjusted the input image resolution of various datasets to match that of the LLVIP dataset. 
\subsection{Ablation Study}
\label{subsec:ablation}

\subsubsection{\textbf{Effectiveness of Each Component}}~To rigorously assess the efficacy of the RSR and DFS modules, we perform an ablation study on each component within RSDet, engaging in an extensive series of experiments on the FLIR, LLVIP, and KAIST dataset, as shown in Table~\ref{tab:ablation}. Both the RSR and DFS modules individually demonstrate consistent improvements across the FLIR, LLVIP, and KAIST datasets. Moreover, the superior results can be obtained by utilizing both RSR and DFS modules, as can be seen from the bold results in Table~\ref{tab:ablation}. These results highlight the effectiveness of the RSR and DFS modules in enhancing the model's detection capabilities.

% 一般现在时不是过去时
\subsubsection{\textbf{Effectiveness of DFS Module}}
To validate the superiority of our proposed DFS method, we replace the DFS module with the other existing RGB-IR fusion module. We have conducted the comparative experiments under a unified backbone and detector. Specifically, we use a two-stream Faster R-CNN as the baseline and apply a direct feature addition fusion method for RGB and IR features~\cite{liu2016multispectral}. For the comparison fusion methods from other studies, we replace the feature addition operation with their own feature fusion module and performed fair experiments on the FLIR dataset. Additionally, we also compare the models in terms of parameter quantity and inference computational complexity.

The experimental results are shown in Table~\ref{tab: fusion method}. Our method increases the number of parameters by only 3.91M compared to the baseline, yet it significantly improves detection performance, with a 7.8\% increase in $\text{mAP}_{50}$, a 4.0\% increase in $\text{mAP}_{75}$, and a 4.2\% increase in mAP. Besides, our DFS module exhibits superior comprehensive performance while maintaining a lower number of parameters and computational complexity. Our method outperforms the second-best fusion method, even with 236.68M fewer parameters and 93G fewer FLOPS, showing a 0.4\% lead in $\text{mAP}_{50}$, a 2.6\% advantage in $\text{mAP}_{75}$, and a 1.6\% improvement in mAP, achieving the best performance among different RGB-IR feature fusion methods.

\begin{table*}[!tbp]
     \centering
    \caption{Detection results~($\text{MR}^{\text{-2}}$, in\%) under \textbf{`All' settings} of different pedestrian distances, occlusion levels, and light conditions (Day and Night) on the KAIST dataset. The pedestrian distances consist of `Near' (115 $\leq$ \emph{height}), `Medium' (45 $\leq$ \emph{height} $<$ 115) and `Far' (1 $\leq$ \emph{height} $<$ 45), while occlusion levels consist of `None' (never occluded), `Partial' (occluded to some extent up to half) and `Heavy' (mostly occluded). IoU $=0.5$ and $0.7$ is set for evaluation. The best results are highlighted in \underline{\textcolor{red}{\textbf{red}}} and the second-place are highlighted in \textcolor{blue!70}{\textbf{blue}}. 
    Noted: We conduct the experiments under the\textbf{`All' setting}, which evaluates the model performance on all objects of the KAIST test dataset, rather than the `Reasonable' setting, which only consists of none/partially occluded objects, and objects larger than 55 pixels.}
    % \vspace{-1.5em}
        \vspace{-0.5em}
    \label{tab:kaist}
    \setlength{\tabcolsep}{3.5mm}
    \renewcommand{\arraystretch}{1.05}

\begin{tabular}{c|c|ccc|ccc|ccc}   
\multicolumn{11}{c}{\textbf{(a) IoU=0.5}} \\
\hline
\rowcolor{lightgray!50} &&\multicolumn{3}{c|}{Scale.}&\multicolumn{3}{c|}{Occlusion.}&\multicolumn{3}{c}{All.}\\ \cline{3-11}
\rowcolor{lightgray!50}\multirow{-2}{*}{Methods} & \multirow{-2}{*}{Backbone}&  near  & medium & far   & none  & partial & heavy  & day   & night & all   \\    \hline
ACF~\cite{hwang2015multispectral}      &   VGG16   & 28.74 & 53.67  & 88.20 & 62.94 & 81.40   & 88.08   & 64.31 & 75.06 & 67.74   \\
Halfway Fusion~\cite{liu2016multispectral}&  VGG16   & 8.13  & 30.34  & 75.70 & 43.13 & 65.21   & 74.36   & 47.58 & 52.35 & 49.18   \\
FusionRPN+BF~\cite{konig2017fully}&   VGG16  & 0.04  & 30.87  & 88.86 & 47.45 & 56.10   & 72.20   & 52.33 & 51.09 & 51.70   \\
IAF R-CNN~\cite{li2019illumination}& VGG16    & 0.96  & 25.54  & 77.84 & 40.17 & 48.40   & 69.76   & 42.46 & 47.70 & 44.23   \\
IATDNN+IASS~\cite{guan2019fusion}  &  VGG16   & 0.04  & 28.55  & 83.42 & 45.43 & 46.25   & 64.57   & 49.02 & 49.37 & 48.96   \\
CIAN~\cite{zhang2019cross}      &  VGG16   & 3.71  & 19.04  & 55.82 & 30.31 & 41.57   & 62.48   & 36.02 & 32.38 & 35.53   \\
MSDS-R-CNN~\cite{li2018multispectral}&  VGG16   & 1.29  & 16.19  & 63.73 & 29.86 & 38.71   & 63.37   & 32.06 & 38.83 & 34.15   \\
AR-CNN~\cite{zhang2019weakly}  &  VGG16   & {0.00}  & 16.08  & 69.00 & 31.40 & 38.63   & \textcolor{blue!70}{\textbf{55.73}}   & 34.36 & 36.12 & 34.95   \\
MBNet~\cite{zhou2020improving}& ResNet50  & {0.00}  & 16.07  & 55.99 & 27.74 & 35.43   & 59.14   & 32.37 & 30.95 & 31.87   \\
% \textcolor{red}{BAANet 2022} ~\cite{yang2022baanet}   & ResNet50  & 0.00   & 13.72  & 51.25 & 25.15 & 34.07  &57.92   & - &  - & -   \\
TSFADet~\cite{yuan2022translation}    & ResNet50  & {0.00}  & 15.99  & 50.71 & 25.63 & 37.29   & 65.67   & 31.76 & 27.44 & 30.74   \\ 
CMPD~\cite{li2022confidence} & ResNet50  & {0.00}   & \textcolor{blue!70}{\textbf{12.99}}  & 51.22 & \textcolor{blue!70}{\textbf{24.04}} & 33.88  &59.37   & \textcolor{blue!70}{\textbf{28.30}} &  30.56 & 28.98   \\
% C²Former-Cas.~\cite{yuan2023mathbf}    & ResNet50  & {0.00}  & 13.71  & \textcolor{blue!70}{\textbf{48.14}} &23.91 & \textcolor{red}{\textbf{32.84}}   & 57.81   & 28.48 & \textcolor{blue!70}{\textbf{26.67}} & \textcolor{blue!70}{\textbf{28.39}}   \\ 
CAGTDet~\cite{yuan2024improving}    & ResNet50  & {0.00}  & 14.00  & \textcolor{blue!70}{\textbf{49.40}} & 24.48 & \underline{\textcolor{red}{\textbf{33.20}}}   & 59.35   & 28.79 & \textcolor{blue!70}{\textbf{27.73}} & \textcolor{blue!70}{\textbf{28.96}}   \\ \hline

% \textcolor{red}{HAFNet~\cite{peng2023hafnet}}        & ResNet50  & \textbf{0.00}   & 13.68  & 53.94 & 26.31 & \textbf{30.10}  &55.16   & 31.11&28.51 &30.49  \\
% \textcolor{red}{ICAFusion~\cite{shen2024icafusion}} &&&&&&&&&&\\\hline
   
% RSDet (Ours)  & ResNet50& \underline{\textcolor{red}{\textbf{0.00}}}  & \underline{\textcolor{red}{\textbf{12.13}}} & \underline{\textcolor{red}{\textbf{39.80}}} & \underline{\textcolor{red}{\textbf{20.49}}} & \textcolor{blue!70}{\textbf{33.25}}   & \textcolor{blue!70}{\textbf{57.60}}   & \underline{\textcolor{red}{\textbf{25.83}}} & \underline{\textcolor{red}{\textbf{26.48}}} & \underline{\textcolor{red}{\textbf{26.02}}}   \\ \hline
RSDet (Ours)  & ResNet50& \underline{\textcolor{red}{\textbf{0.00}}}  & \underline{\textcolor{red}{\textbf{10.69}}} & \underline{\textcolor{red}{\textbf{37.68}}} & \underline{\textcolor{red}{\textbf{18.97}}} & \textcolor{blue!70}{\textbf{33.27}}   & \underline{\textcolor{red}{\textbf{55.59}}}  & \underline{\textcolor{red}{\textbf{24.18}}} & \underline{\textcolor{red}{\textbf{26.49}}} & \underline{\textcolor{red}{\textbf{24.79}}}   \\ \hline

\multicolumn{11}{c}{} \\
\multicolumn{11}{c}{\textbf{(b) IoU=0.7}} \\
% \begin{tabular}{c|c|cccccc|ccc}    
\hline
\rowcolor{lightgray!50} &&\multicolumn{3}{c|}{Scale.}&\multicolumn{3}{c|}{Occlusion.}&\multicolumn{3}{c}{All.}\\ \cline{3-11}
\rowcolor{lightgray!50}\multirow{-2}{*}{Methods} & \multirow{-2}{*}{Backbone}&  Near  & Medium & Far   & None  & Partial & Heavy  & Day   & Night & All   \\    \hline

ACF~\cite{hwang2015multispectral}      &   VGG16    & 79.25 & 82.96 & 97.86 & 87.59 & 94.61 & 97.86 & 88.48 & 92.47 & 89.54    \\
Halfway Fusion~\cite{liu2016multispectral}&  VGG16   & 49.59 & 74.87 & 97.00   & 80.35 & 90.42 & 94.15 & 81.31 & 86.34 & 83.15   \\
FusionRPN+BF~\cite{konig2017fully}&   VGG16  & 35.78 & 68.82 & 99.38 & 76.29 & 86.80  & 92.47 & 76.98 & 83.71 & 79.30   \\
IAF R-CNN~\cite{li2019illumination}& VGG16    & 33.75 & 70.24 & 98.12 & 76.74 & 84.58 & 93.69 & 77.02 & 84.38 & 79.59   \\
IATDNN+IASS~\cite{guan2019fusion}  &  VGG16   & 45.40  & 70.85 & 99.00    & 78.25 & 84.51 & 93.13 & 80.46 & 82.32 & 80.91   \\
CIAN~\cite{zhang2019cross}      &  VGG16   & 38.31 & 63.98 & 87.12 & 70.39 & 80.95 & 91.68 & 72.44 & 78.92 & 74.45   \\
MSDS-R-CNN~\cite{li2018multispectral}&  VGG16   & 35.49 & 57.95 & 93.15 & 68.41 & 76.23 & 90.37 & 69.85 & 78.52 & 71.93   \\
AR-CNN~\cite{zhang2019weakly}  &  VGG16   & 25.19 & 53.88 & 91.72 & 64.91 & 73.18 & 88.70  & 64.45 & 77.29 & 68.64   \\
MBNet~\cite{zhou2020improving}& ResNet50  & \textcolor{blue!70}{\textbf{16.98}} & 51.21 & 85.33 & 60.84 & 69.59 & 86.22 & 63.50  & 67.76 & 65.14   \\
% \textcolor{red}{BAANet 2022} ~\cite{yang2022baanet}   & ResNet50  & 0.00   & 13.72  & 51.25 & 25.15 & 34.07  &57.92   & - &  - & -   \\
TSFADet~\cite{yuan2022translation}    & ResNet50  & 19.50  & 49.32 & 81.90  & 58.93 & 72.09 & 87.10  & 61.78 & 68.38 & 63.85   \\ 
CMPD~\cite{li2022confidence} & ResNet50  & 19.31 & 49.69 & 83.93 & 59.79 & \textcolor{blue!70}{\textbf{66.64}} & \textcolor{blue!70}{\textbf{84.79}} & 61.77 & 68.83 & 63.93   \\
% C²Former-Cas.~\cite{yuan2023mathbf}    & ResNet50  & {0.00}  & 13.71  & \textcolor{blue!70}{\textbf{48.14}} &23.91 & \textcolor{red}{\textbf{32.84}}   & 57.81   & 28.48 & \textcolor{blue!70}{\textbf{26.67}} & \textcolor{blue!70}{\textbf{28.39}}   \\ 
CAGTDet~\cite{yuan2024improving}    & ResNet50  & 20.80  & \textcolor{blue!70}{\textbf{47.40}}  & \textcolor{blue!70}{\textbf{78.31}} & \textcolor{blue!70}{\textbf{56.95}} & 67.39 & 85.11 & \textcolor{blue!70}{\textbf{60.24}} & \textcolor{blue!70}{\textbf{65.45}} & \textcolor{blue!70}{\textbf{61.71}}  \\ \hline

% \textcolor{red}{HAFNet~\cite{peng2023hafnet}}        & ResNet50  & \textbf{0.00}   & 13.68  & 53.94 & 26.31 & \textbf{30.10}  &55.16   & 31.11&28.51 &30.49  \\
% \textcolor{red}{ICAFusion~\cite{shen2024icafusion}} &&&&&&&&&&\\\hline

% RSDet (Ours)  & ResNet50& \underline{\textcolor{red}{\textbf{12.68}}} & \underline{\textcolor{red}{\textbf{46.73}}} & \underline{\textcolor{red}{\textbf{73.95}}} & \underline{\textcolor{red}{\textbf{55.35}}} & \underline{\textcolor{red}{\textbf{65.03}}} & \underline{\textcolor{red}{\textbf{84.08}}} & \underline{\textcolor{red}{\textbf{59.72}}} & \underline{\textcolor{red}{\textbf{62.46}}} & \underline{\textcolor{red}{\textbf{60.65}}} \\ \hline
RSDet (Ours)  & ResNet50& \underline{\textcolor{red}{\textbf{11.81}}} & \underline{\textcolor{red}{\textbf{46.03}}} & \underline{\textcolor{red}{\textbf{71.62}}} & \underline{\textcolor{red}{\textbf{54.41}}} & \underline{\textcolor{red}{\textbf{65.79}}} & \underline{\textcolor{red}{\textbf{82.24}}} & \underline{\textcolor{red}{\textbf{59.17}}} & \underline{\textcolor{red}{\textbf{62.54}}} & \underline{\textcolor{red}{\textbf{60.04}}} \\ \hline
\end{tabular}
\end{table*}

\subsubsection{\textbf{Comparative Analysis of Filter Designs within the RSR Module}}
We further explore the impact of different Filter Type designs and different values of $K$ in the RSR module.
Specifically, if the positions of $M_{m_I}$ and $M_{m_V}$ not corresponding to the $\operatorname{top}K$ values are in the range of $\left[0, 1\right]$, the Filter is the soft filter, And if are directly set to 0 is the hard Filter. 
We have conducted the experiments by adjusting the value of $K$ from 300 to 400 across different filter types on the FLIR datasets. Note that 400 is the total patch number of image, so, $K=400$ represents nothing removed.
As illustrated in Table~\ref{tab: design rsr}, our detector achieves the highest detection performance with the Soft Filter when $K=320$. In addition to considering the optimal performance, we also evaluate the stability of each filter type. To this end, we calculate the average (Avg.) mAP value across different $K$ for each filter type. From Table~\ref{tab: design rsr}, on the FLIR-aligned dataset, it is evident that the Soft Filter consistently outperforms the others across all metrics, with the best result also achieved using the Soft Filter. Therefore, we select the Soft Filter and $K=320$ as the filter hyper-parameters.

\begin{table}[!t]
\begin{center}
    \caption{Camparison the performance (mAP, in\%) on the FLIR dataset. The best results are highlighted in \underline{\textcolor{red}{\textbf{red}}} and the second-place are highlighted in \textcolor{blue!70}{\textbf{blue}}.}
        % \vspace{0.2em}
    % \vspace{-0.5em}
    \label{tab:flir}
    \setlength{\tabcolsep}{1.5mm}
        \renewcommand{\arraystretch}{1.0}
    
    \begin{tabular}{cc|cc|c}
    \hline
\rowcolor{lightgray!50}&&\multicolumn{2}{c|}{FLIR}& \\ 
\rowcolor{lightgray!50}\multirow{-2}{*}{Methods} & \multirow{-2}{*}{Backbone}&$\text{mAP}_{50}\uparrow$&$\text{mAP}\uparrow$  & \multirow{-2}{*}{Modality} \\     \hline
  
    SSD~\cite{liu2016ssd}  & VGG16&65.5&29.6  &  \\
    RetinaNet \cite{lin2017focal}&ResNet50&66.1&31.5& \\
        % Mask R-CNN &&&&&& \\
    Cascade R-CNN~\cite{cai2019cascade} &ResNet50&71.0&34.7& \\
     Faster R-CNN~\cite{ren2017faster}& ResNet50 &74.4& 37.6   & \\
   
     DDQ-DETR~\cite{zhang2023dense}&ResNet50 &73.9&37.1 &\multirow{-5}{*}{IR}
   \\\hline \hline
        SSD~\cite{liu2016ssd}  & VGG16 & 52.2&21.8  &  \\
        RetinaNet~\cite{lin2017focal} &ResNet50&51.2&21.9& \\
        Cascade R-CNN~\cite{cai2019cascade}&ResNet50&56.0&24.7& \\
         Faster R-CNN~\cite{ren2017faster}& ResNet50 &64.9&28.9  & \\
       DDQ-DETR~\cite{zhang2023dense} &ResNet50 &64.9&30.9       &\multirow{-5}{*}{RGB}\\
       \hline \hline

    Halfway fusion~\cite{liu2016multispectral}& VGG16&71.5&35.8 &    \\ 
    CFR\_3~\cite{zhang2020multispectral}&VGG16 &72.4&  - & \\
    GAFF~\cite{zhang2021guided}&VGG16&72.7&37.3 &  \\
       GAFF~\cite{zhang2021guided}&ResNet18&74.6&37.4  &   \\
    CAPTM\_3~\cite{zhou2021visible}&ResNet50&73.2&-   &\\
    CMPD~\cite{li2022confidence}&ResNet50 &69.4&-  &  \\

    LGADet~\cite{zuo2023lgadet} &ResNet50&74.5&- & \\
    ProbEn~\cite{chen2022multimodal}& ResNet50&75.5&37.9 & \\
    TINet~\cite{zhang2023illumination} &ResNet50&76.1&36.5 & \\
    ICAFusion~\cite{shen2024icafusion}&ResNet50&72.0& &\\ 
    ICAFusion~\cite{shen2024icafusion}&CSPDarkNet53&  79.2&41.4 &\\
    % CFT~\cite{qingyun2021cross}&CSPDarknet53&78.7&40.2   \\
    % % CFT~\cite{qingyun2021cross}&Arxiv 2021&Darknet53&&76.3&38.3   \\

    % CFT~\cite{qingyun2021cross}&ResNet50&77.5&39.2  & \\
    YOLOFusion~\cite{qingyun2022cross}& CSPDarkNet53 &76.6&39.8 & \\
    MFPT\cite{zhu2023multi}&ResNet50&80.0 &- \\

    CSAA~\cite{cao2023multimodal}&ResNet50  &\textcolor{blue!70}{\textbf{79.2}}&\textcolor{blue!70}{\textbf{41.3}}   &\\

% RSDet (Ours)&ResNet50&\underline{\textcolor{red}{\textbf{81.1}}} & \underline{\textcolor{red}{\textbf{41.4}}}  &\multirow{-15}{*}{RGB+IR} \\     
RSDet~(Ours)&ResNet50&\underline{\textcolor{red}{\textbf{83.9}}} & \underline{\textcolor{red}{\textbf{43.8}}}  &\multirow{-15}{*}{RGB+IR} \\     
  \hline
    \end{tabular}
       \vspace{-1.5em}
   % \vspace{0.6em}
\end{center}
\end{table}

\subsection{Visualization of Intermediate Results}

\subsubsection{\textbf{Visualizations of the intermediate results in the RSR module}}
To demonstrate the effectiveness of the RSR module, we visualize the intermediate results, as illustrated in Figure~\textcolor{red}{\ref{fig:rsr}}.
It is evident that the object regions remain largely unchanged after processing through the RSR module, while the removed information is predominantly concentrated in the background areas. This indicates that the RSR module adaptively filters out redundant background noise that is irrelevant to the detection task. For a more objective comparison, we calculate the signal-to-noise ratio (SNR) to quantify the differences. The results reveal a modest increase in SNR after the original image passes through the RSR module, further substantiating the efficacy of the RSR module.

\subsubsection{\textbf{Visualizations of the Shared, Specific and Fused Features}}

We provide visualizations of the shared and specific features, denoted as $C_{\text{sha}}$, $C_{\text{I-spe}}$, and $C_{\text{V-spe}}$, as depicted in Figure~\ref{fig: DFS_vis}. In these visualizations, red boxes are used to highlight the objects of interest. By examining the features before and after the fusion process, we can clearly see the impact of our method. Specifically, the DFS module enhances the visibility of non-salient objects within the shared feature space, transforming them into salient features that are more distinguishable. This indicates that the fusion method effectively integrates information from different modalities, leading to a more comprehensive representation.
\subsubsection{\textbf{tSNE Visualizations of the RSR module}}

~We also conduct tSNE visualizations of shared ~($C_{\text{sha}}$) and specific features~($C_{\text{I-spe}}$ and $C_{\text{V-spe}}$) on the FLIR dataset. As shown in Figure~\ref{fig: rsr_tsne}, it can be observed that without the RSR module, there is still a considerable mix of feature points in the shared and specific features, leading to difficulty in selecting the desired specific features in DFS. After adding the RSR module, we have noted a significant decrease in the number of mixed features, which verifies that removing redundant spectra is beneficial to feature disentanglement and, thus more effective in obtaining fused features in the DFS module.

\begin{table}[!t]
\begin{center}
    \caption{Camparison the performance (mAP, in\%) on the LLVIP dataset. The best results are highlighted in \underline{\textcolor{red}{\textbf{red}}} and the second-place are highlighted in \textcolor{blue!70}{\textbf{blue}}.}
       % \vspace{-0.5em}
    \label{tab:llvip}
    \setlength{\tabcolsep}{1.5mm}
        \renewcommand{\arraystretch}{1.1}
    
    \begin{tabular}{cc|cc|c}
    \hline
\rowcolor{lightgray!50}&&\multicolumn{2}{c|}{LLVIP} & \\ 
\rowcolor{lightgray!50}\multirow{-2}{*}{Methods} & \multirow{-2}{*}{Backbone}&$\text{mAP}_{50}\uparrow$&$\text{mAP}\uparrow$  & \multirow{-2}{*}{Modality} \\     \hline
  
    SSD~\cite{liu2016ssd}  & VGG16&90.2&53.5 &  \\
    RetinaNet \cite{lin2017focal}&ResNet50&94.8&55.1& \\
        % Mask R-CNN &&&&&& \\
    Cascade R-CNN~\cite{cai2019cascade} &ResNet50&95.0&56.8 & \\
     Faster R-CNN~\cite{ren2017faster}& ResNet50 & 94.6 & 54.5& \\
   
     DDQ-DETR~\cite{zhang2023dense}&ResNet50 &93.9&58.6&\multirow{-5}{*}{IR}
   \\\hline \hline
        SSD~\cite{liu2016ssd}  & VGG16  &82.6&39.8 &  \\
        RetinaNet~\cite{lin2017focal} &ResNet50&88.0&42.8& \\
        Cascade R-CNN~\cite{cai2019cascade}&ResNet50&88.3&47.0& \\
         Faster R-CNN~\cite{ren2017faster}& ResNet50 & 87.0 & 45.1 & \\
       DDQ-DETR~\cite{zhang2023dense} &ResNet50 &86.1&46.7
       &\multirow{-5}{*}{RGB}\\
       \hline \hline
    Halfway fusion~\cite{liu2016multispectral}& VGG16&91.4  & 55.1 &    \\ 
    GAFF~\cite{zhang2021guided}&ResNet18& 94.0&   55.8 &   \\
      ProbEn~\cite{chen2022multimodal}& ResNet50 &  93.4  & 51.5 & \\
    CSAA~\cite{cao2023multimodal}&ResNet50   &\textcolor{blue!70}{\textbf{94.3}} &\textcolor{blue!70}{\textbf{59.2}}  &\\

RSDet~(Ours)&ResNet50&\underline{\textcolor{red}{\textbf{95.8}}}&\underline{\textcolor{red}{\textbf{61.3}}}  &\multirow{-5}{*}{RGB+IR} \\     
  \hline
    \end{tabular}
   \vspace{-1.5em}
\end{center}
\end{table}

\subsection{Comparison with State-of-the-Art Methods}

\subsubsection{\textbf{Comparision on the KAIST Datas}}
\label{subsubsec:experiment_comp_kaist}

We compare our proposed RSDet with twelve state-of-the-art methods, on the KAIST dataset.
Table~\ref{tab:kaist} provides the performance comparisons, we calculate the $\text{MR}^{\text{-2}}$ under IoU thresholds of 0.5 and 0.7. It's worth noting that the "All" setting is more challenging, as it includes small~(less than 55 pixels) and heavily occluded objects. Our method achieves state-of-the-art results in the "All" setting, indicating not only strong overall detection performance but also effectiveness at detecting small and heavy occlusion objects.

When IoU = 0.5, according to Table~\ref{tab:kaist}\textcolor{red}{(a)}, RSDet demonstrates outstanding performance, leading in the "ALL" setting (including all', day', and night') and dominating five of the six subsets (near', medium', far', none', heavy'). It ranks second only on the partial' subset. Notably, RSDet exhibits a significant advantage in subsets of different scales (None', Medium', Far'), especially in the Far' subset, where it outperforms the second-best by an impressive margin of approximately 11.72\%. We infer that this superior performance can be attributed to the mixture of scale-aware experts in the DFS module, which significantly enhances the model's ability to perceive objects at different distances. Moreover, across the entire test dataset (`all'), RSDet surpasses the second-best 4.17\%, underscoring its overall superiority. 

When IoU=0.7, RSDet also performs best across all subsets, as shown in Table~\ref{tab:kaist}\textcolor{red}{(b)}. As the IoU increases, the requirement for detection accuracy becomes more stringent. In this scenario, the superiority of our method becomes more apparent.
For example, all the recent methods have an $\text{MR}^{\text{-2}}$ of 0 under the 'near' subset before the IoU increases to 0.7. However, These methods show an increase in Miss Rate ranging from a maximum of 25.19\% to a minimum of 16.98\%, whereas our method only rises by 11.81\%, the $\text{MR}^{\text{-2}}$ of RSDet is significantly better than other methods, indicating that our proposed method produces more accurate detection results. 

\begin{figure*}[!t]
    \centering
    \includegraphics[width=0.9\linewidth]{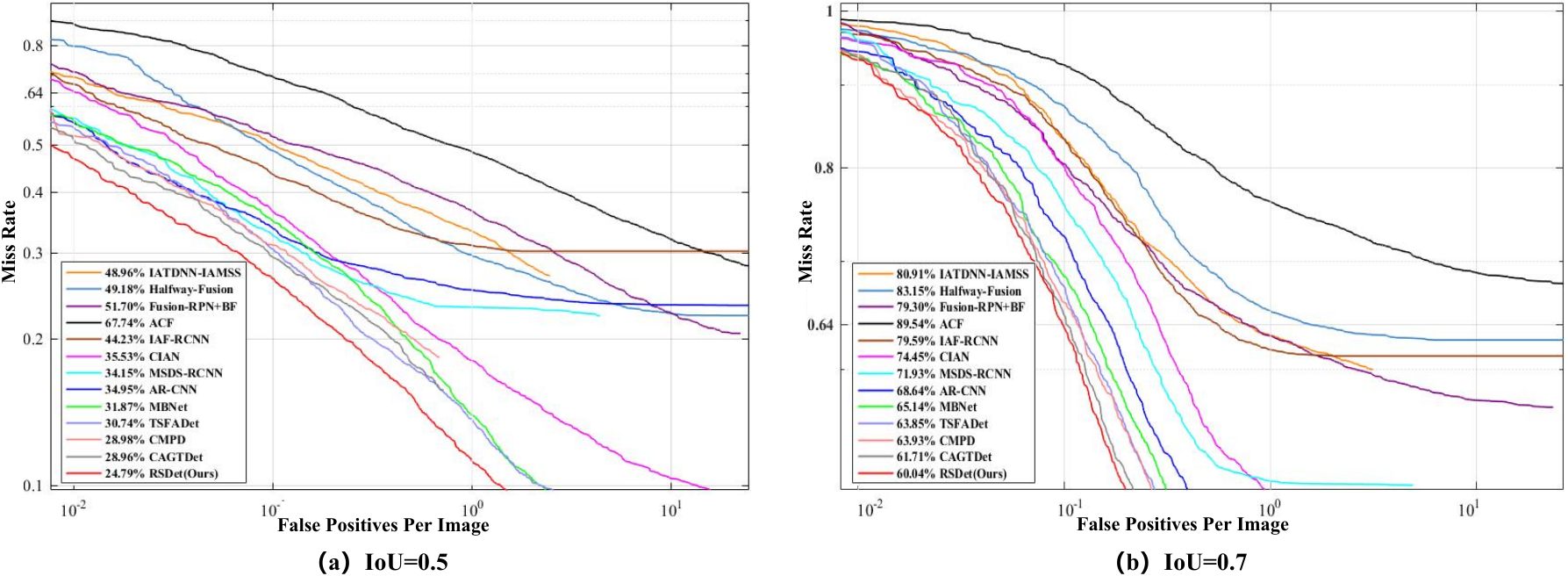}
        \vspace{-0.5em}
    \caption{The MR-FPPI curves comparisons with the state-of-the-art methods on the KAIST dataset under the `All' settings.}
    % \vspace{-0.5em}
    % \label{fig2}
    \vspace{-1.5em}
    \label{fig: kaist}
\end{figure*}

To conduct a more comprehensive analysis, we illustrate the Log-average Miss Rate over the False Positive Per Image (FPPI) curve in Figure~\ref{fig: kaist}, which highlights the superiority of our method, primarily reflected in the following points: 
1)~Significant Miss Rate Advantage: RSDet achieves the lowest miss rate among all methods, indicating a stronger ability to identify objects accurately, even under more strict conditions. 2)~Excellent False Positive Control: RSDet maintains a notably lower MR even at lower FPPI, demonstrating its effectiveness in reducing false detection while maintaining high detection precision. This enhances the reliability of the detection system. 
3)~Smooth Curve Performance: The MR-FPPI curve of RSDet is smoother as FPPI increases. This suggests that RSDet offers stable performance across various conditions, making it adaptable to diverse scenarios.

\subsubsection{\textbf{Comparision on the FLIR Dataset}}The quantitative results of the different methods on the FLIR datasets are shown in Table~\ref{tab:flir}. We compare the proposed RSDet with twelve SOTA RGB-IR object detection methods on the FLIR dataset.
From Table~\ref{tab:flir}, it can be seen that the fusion methods generally outperform the single-modal methods on the FLIR dataset. Our RSDet method also stands out, achieving 83.9\% $\text{mAP}_{50}$ and 43.8\% mAP, significantly surpassing the second-best RGB-IR methods by 4.7\% and 2.5\%, respectively, demonstrating the superior ability of our method on the FLIR Dataset.

\subsubsection{\textbf{Comparision on the LLVIP Dataset}}
The quantitative results of the different methods on the LLVIP datasets are shown in Table~\ref{tab:llvip}. Due to the LLVIP dataset having only recently been made public, there are relatively few published methods that have conducted experiments on it. Thus, we compare with four SOTA RGB-IR methods which have been compared on FLIR comparison experiments, and the five single-modality methods. In the LLVIP dataset, the poor light conditions result in the RGB features interfering with the IR features during the multimodal feature fusion process, leading to the detection results of the RGB-IR detection method always being outperformed by the single IR modality method. Our method effectively addresses this issue and achieves 95.8\% $\text{mAP}_{50}$ and 61.3\% mAP, surpassing the second-best RGB-IR methods by 1.2\% and 2.1\%, respectively. The adequate comparison experiments verify the effectiveness of our coarse-to-fine fusion strategy and achieve state-of-the-art performance on the KAIST, FLIR, and LLVIP datasets.

\section{Conclusion}
\label{sec: Conclusion}
In this paper, we presented a new coarse-to-fine perspective to fuse visible and infrared modality features. Specifically, a Redundant Spectrum Removal (RSR) module is first designed to coarsely filter out the irrelevant spectrum, and then a Dynamic Feature Selection (DFS) module is proposed to finely select the desired features for the RGB-IR final feature fusion process. we constructed a new object detector called Removal and Selection Detector~(RSDet) to evaluate its effectiveness and versatility. Extensive experiments on three public RGB-IR detection datasets demonstrated that our method can effectively facilitate complementary fusion and achieve state-of-the-art performance. We believe that our method can be applied to various studies in the RGB-IR feature fusion tasks.

% \section{Discusstions and Future Works}
% \begin{itemize}
%     \item LLVIP yolo has a better performance
%     \item trade-off%从整体上来看，我们的corase-to-fine fusion stratgy提升了map但是，但是这里出现了一个trade-off 以牺牲很少map75来换取了更高的map50，具体表现为map75通常会出现相应的下降
%     \item reasonable subset on the KAIST dataset
%     \item 涉及傅里叶变换，RSR模块时间效率低
%       \item  data augmentation使用少，性能仍有很大提升空间。
% \end{itemize}
\label{sec: Future Works}

{
    \small
    \bibliographystyle{IEEEtran}
    \bibliography{tmm}
}

\newpage

\end{document}